\documentclass{article}
\usepackage{spconf,amsmath,graphicx}
\usepackage{subfigure}
\usepackage{here}

\usepackage{siunitx} 
\usepackage{color}

\newcommand{\fnoteII}[1]{{\color{magenta} \bf #1 \color{black}}}
\newcommand{\knote}[1]{{\color{red} \bf #1 \color{black}}} 
\newcommand{\kcutcandidate}[1]{{\color{green} \bf #1 \color{black}}}
\newcommand{\bnote}[1]{{\color{red} #1 \color{black}}}
\newcommand{\review}[1]{{\color{blue} \bf #1 \color{black}}}
\newcommand{\onote}[1]{\color{blue} #1 \color{black}}

\newcommand{\kcut}[1]{}
\newcommand{\scut}[1]{}
\newcommand{\ocut}[1]{}
\newcommand{\jptext}[1]{{\color{blue} \bf #1 \color{black}}}
\ifx\pdfoutput\undefined
\else
 \usepackage{CJKutf8}
\renewcommand{\knote}[1]{}
 \renewcommand{\fnoteII}[1]{}
 \renewcommand{\review}[1]{}
 \renewcommand{\kcutcandidate}[1]{}
 \renewcommand{\bnote}[1]{}
 \renewcommand{\onote}[1]{}
 \renewcommand{\jptext}[1]{}
\fi

\ifx\argmin\undefined\newcommand{\argmin}{\mathop{\rm argmin}\limits}\fi
\ifx\argmax\undefined\newcommand{\argmax}{\mathop{\rm argmax}\limits}\fi
\ifx\bm\undefined\newcommand{\bm}[1]{\mbox{\boldmath{$#1$}}}\fi
\ifx\um\undefined\newcommand{\um}[1]{{\SI{#1}{\micro \metre}}}\fi 

\ifx\etal\undefined\newcommand{\etal}{{\it et al. }}\fi
\ifx\ie\undefined\newcommand{\ie}{{\it i.e.}}\fi
\ifx\eg\undefined\newcommand{\eg}{{\it e.g.}}\fi
\ifx\gt\undefined\newcommand{\gt}{$\textgreater$}\fi
\ifx\lt\undefined\newcommand{\lt}{$textless$}\fi

\graphicspath{{figs/}{../figs/}{../}}

\newcommand{\cardboard}{{cardboard}}
\newcommand{\shoes}{{shoes}}
\newcommand{\mesh}{{mesh}}
\newcommand{\basket}{{basket}}

\newcommand{\fref}[1]{Fig. \ref{#1}}

\title{High-frequency shape recovery from shading by \\ CNN and domain adaptation}
\name{Kodai Tokieda, Takafumi Iwaguchi, Hiroshi Kawasaki}
\address{Information Science and Electrical Engineering, Kyushu University, Fukuoka, Japan}
\begin{document}
%
\maketitle
\begin{abstract}
Importance of structured-light based one-shot scanning technique is increasing because of its simple system configuration and ability of capturing moving objects.
One severe limitation of the technique is that it can capture only sparse shape, but not high frequency shapes, because certain area of projection pattern is required to encode spatial information.
In this paper, we propose a technique to recover high-frequency shapes by using shading information, which is captured by one-shot RGB-D sensor based on structured light with single camera.
Since color image comprises shading information of object surface, high-frequency shapes can be recovered by shape from shading techniques.
Although multiple images with different lighting positions are required for shape from shading techniques, we propose a learning based approach to recover shape from a single image.
In addition, to overcome the problem of preparing sufficient amount of data for training, we propose a new data augmentation method for high-frequency shapes using synthetic data and domain adaptation.
Experimental results are shown to confirm the effectiveness of the proposed method.
\end{abstract}
\section{Introduction} \label{sec:intro}
The importance of active stereo method based on structured-light is 
increasing because of its simple system configuration and potential of 
high accuracy shape reconstruction. Gray-code and phase-shift, which are known 
as temporal coding methods, are typical implementation and 
have been widely used so far. Since those techniques
require multiple patterns to identify the
pixel location, fast moving object cannot be captured.
To compensate the problem, one-shot scanning methods draw a wide attention, which is based on sparse structured light comprising
codes for identification in the single pattern, achieving shape reconstruction 
from one or only a few pattern projection.  
A common problem of the sparse structured light methods
is that the resolution of reconstruction is much lower than temporal coding 
method, specifically they can only recover the shape around the pattern region, 
which is sparse. 

In this paper, we tackle with this problem to recover dense and high-frequency 
shapes by using shading information, which is captured by 
oneshot RGB-D sensor based on structured light and single camera.
In previous methods, which usually employ
interpolation by median filter, Gabor filter or moving average, the regions which have no pattern projection are reconstructed as a low-frequency shapes, 
i.e., smooth/flat surfaces, even if the surface has high frequency shapes.
With our technique, {\it shading information} of
such no pattern projected regions is analyzed and high-frequency shapes are 
recovered from one-shot (single) color image. 
Since multiple images 
with different lighting positions are usually required for shape from shading 
techniques, it is theoretically impossible to recover shape only from single image. For 
solution, we propose a learning based approach to recover the shapes up to scale 
from 
single image. In addition, to overcome the problem of preparing 
sufficient amount of data for training, we propose a domain adaptation method 
where new training data set are generated by computer graphics technique
using parametric representation of high-frequency shapes. 
Those synthesized shapes are efficiently used to learn high-frequency shapes and a few real data are used to adapt our model to real environment.
Experimental results are shown to 
confirm the effectiveness of the proposed method compared  with previous 
method, e.g., shape super-resolution techniques. 
Our contributions are threefold:

\begin{itemize}
\vspace{-2mm}
  \setlength{\parskip}{0cm} 
  \setlength{\itemsep}{0cm} 

\item Dense and high-frequency shape reconstruction method
using convolutional neural network (CNN), whose input is
a single shading image and a single sparse depth image, is proposed.
\item Efficient domain adaptation 
 method using huge amount of computationally generated images based on parametric 
representation of high-frequency shapes is proposed, which is further improved by 
		  fine tuning technique using real data.
\item It is shown that the proposed network is able to reconstruct dense and 
high-frequency shapes using real data set, comparing with previous methods.
\end{itemize}

\section{Related work} \label{sec:related}
Since Horn proposed shape from shading~\cite{Horn1974, Zhang1999}, methods for recovering object shapes from shading have been proposed such as photometric stereo~\cite{Woodham1980}.
While there are cases where the shape cannot be uniquely determined from the shading~\cite{belhumeur1999bas, Yuille1993}, methods have been proposed to estimate the shape from a single image by using multiple constraints~\cite{Barron2015} or by learning~\cite{Yang2018, Henderson2019}.
The use of high-resolution RGB images has been studied as a clue to increase the resolution of low-resolution shapes obtained by RGB-D cameras.
The method to generate a high-resolution distance image from low-resolution distance images and high-resolution RGB images is called guided super-resolution~\cite{Barron2016, Lutio2019, Lu_2015_CVPR}.
CNN based depth map super-resolution is proposed by ~\cite{Hui16, Chen_2018_ECCV}.
Network training with synthetic data is seen in dense 3D reconstruction~\cite{Richter2016}, and in shape from shading~\cite{Richter2015, Ikehata2018}, where large real datasets are difficult to construct.
Models trained on synthetic data often cannot generalize to real-world data.
As a method of domain adaptation, source to target adversarial learning ~\cite{Tzeng2017, Tsai2018, Hoffman2018} and matching the mean and variance of feature vectors have been proposed~\cite{Sun2016}.



\section{Shape recovery from shading information} \label{sec:recovery}

\subsection{Algorithm overview}

\begin{figure}[t]
    \centering
    \vspace{-5mm}
   \includegraphics[width=0.8\linewidth]{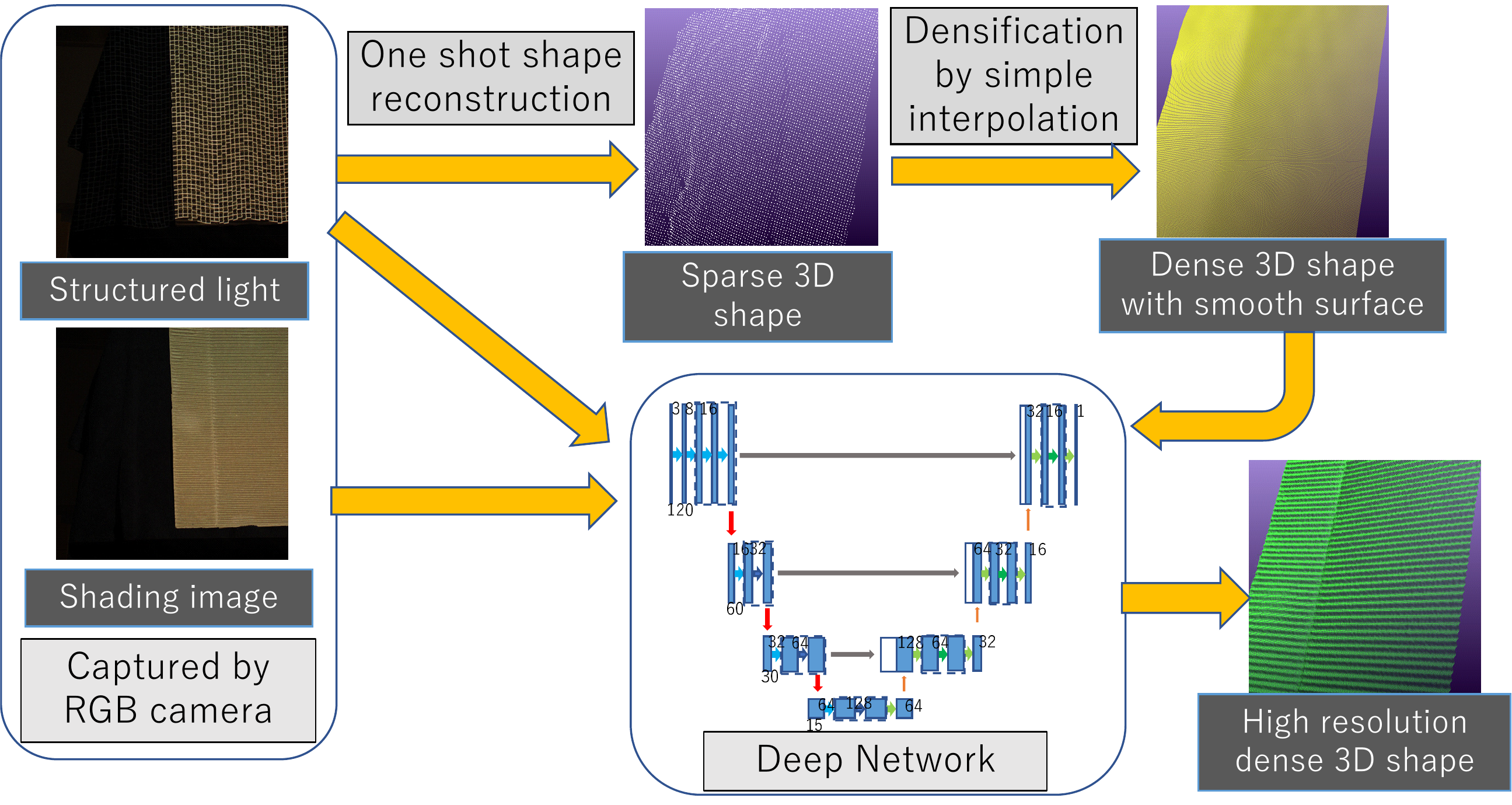}
\vspace{-3mm}
    \caption{Algorithm overview.}
    \vspace{-5mm}
    \label{fig:algorithm}
\end{figure}

Overview of our algorithm is shown in \fref{fig:algorithm}.
Our method reconstructs the high-frequency shape by adding a high-frequency component estimated from shading to the measured low-frequency shape.
First, we measure the low-frequency shape of the object and the shading image under fixed illumination.
For low-frequency shapes, a one-shot measurement with a projector and camera system~\cite{Furukawa2016} is used, but the shape measurement method in our framework is not limited to this technique.
Next, the low-frequency shape and shading images are input into the CNN to reconstruct the high-frequency shape.
Learning the shading of high-frequency shapes requires a lot of training data, however the measurement of many variations of objects is time consuming and the measurement of high-frequency shapes that are ground truth is also difficult.
Therefore, we construct a synthetic dataset of shading images by computer graphics and perform domain adaptation.
What we proposed in this paper is how to deal the lack of training data by data augmentation using appropriate synthetic data and adapt to real data by fine-tuning, we call this whole process domain adaption.

\subsection{System configuration and initial data acquisition}

\begin{figure}
\vspace{-5mm}
  \begin{tabular}{ccc}
    \centering
    \includegraphics[height=0.25\linewidth]{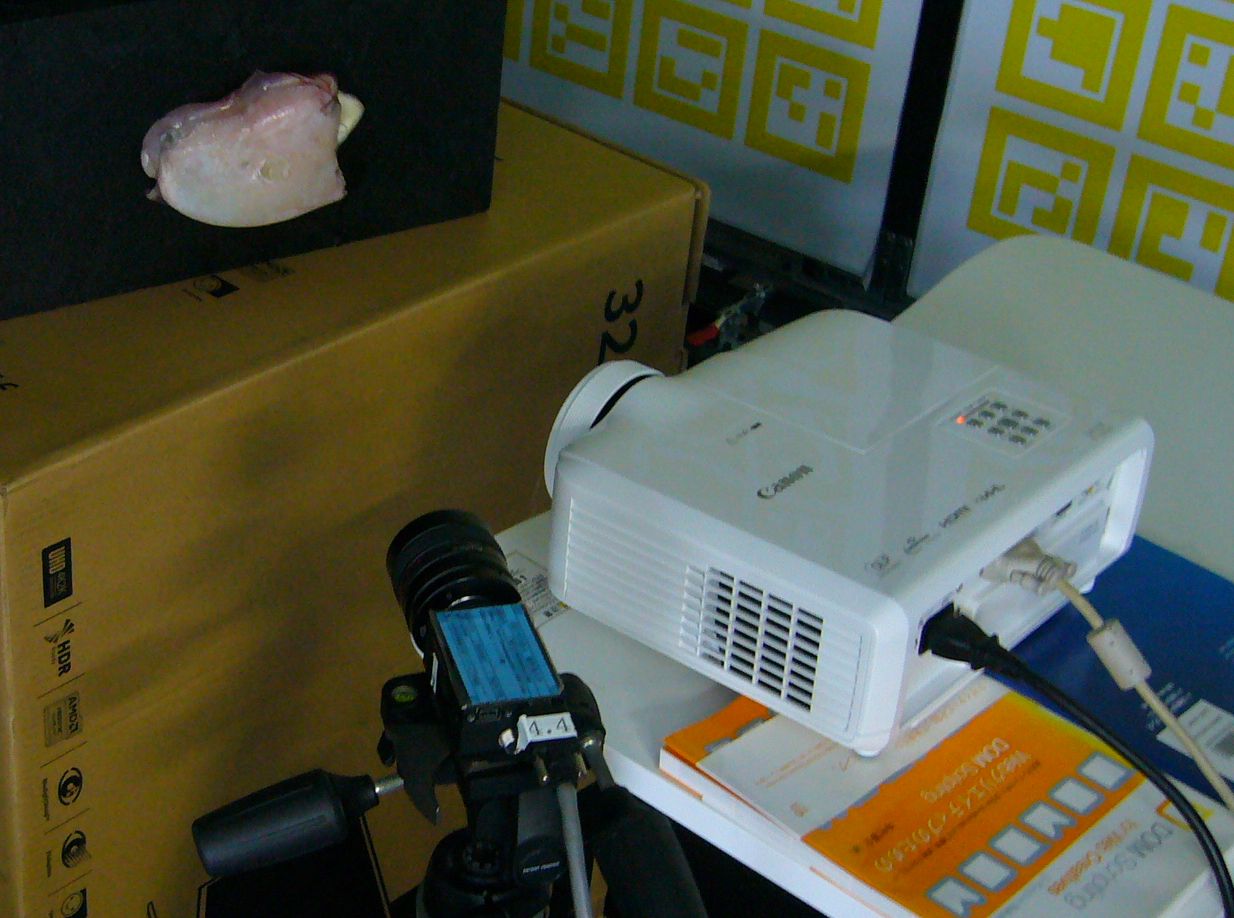}&
    \includegraphics[height=0.25\linewidth]{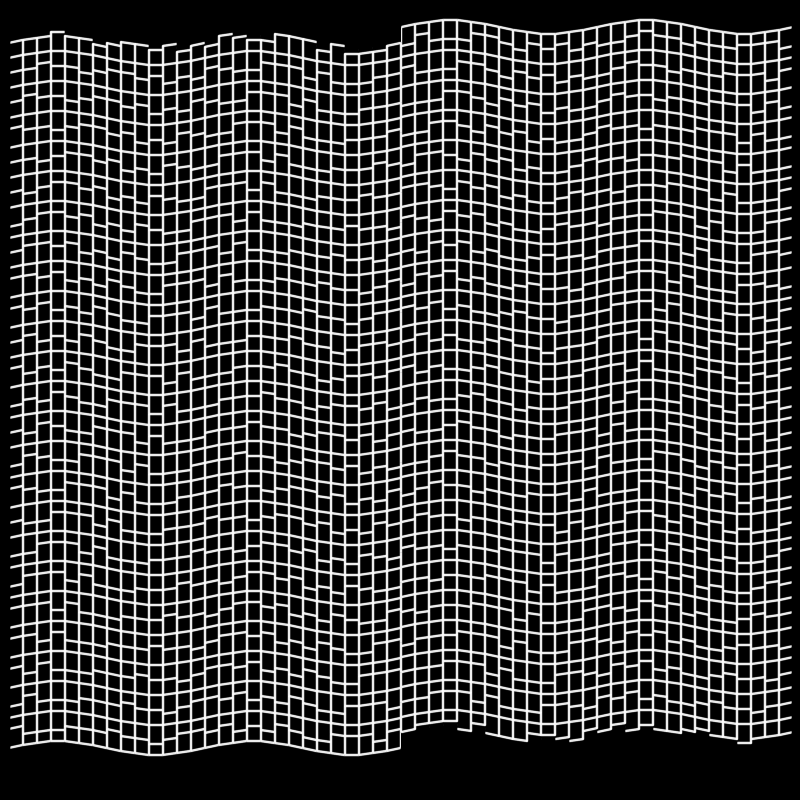}&
    \includegraphics[height=0.25\linewidth]{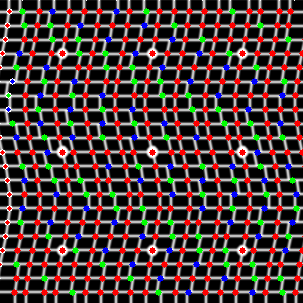}\\
(a) & (b) & (c) \\
  \end{tabular}
    \vspace{-3mm}
    \caption{(a) Setup of the measurement. (b) To measure the low-frequency shape, a sparse grid pattern is projected onto the object. (c) The grid points are decoded into different codes depending on their connection to their neighbors.}
    \label{fig:gapreconst}
\vspace{-5mm}
\end{figure}

One-shot measurement using a projector and camera system with a grid pattern~\cite{Furukawa2016} is used as a method to measure the low-frequency shape.
Setup is shown in \fref{fig:gapreconst}(a).
The pattern projected by the projector consists of a sparse grid pattern (\fref{fig:gapreconst}(b)), with codes embedded as a positional relationship between adjacent grid points (\fref{fig:gapreconst}(c)).
By identifying grid points using codes, the corresponding points to the image captured by the camera are detected, and the distances are calculated by triangulation.
With the grid pattern, only the shapes on the lines that make up the pattern can be measured.
We obtain low-frequency shapes by interpolating the sparse shapes with the radial basis function.


\section{Implementation} 

\subsection{High-frequency shape recovery from shading information using CNN}
We recover the high-frequency shape by shape from shading approach.
Since the shading does not imply an absolute depth, only a relative depth 
can be estimated, and thus, we added estimated high-frequency depth to the measured absolute depth. 
We found that there was an ambiguity on shape reconstruction, such as convex or 
concave, from the same input image. To avoid the ambiguity, we limit the light 
source direction on one side. Note that even after such limitation, the possible 
light source position is still large and it does not decrease the flexible of our method.
While this number of shading images is insufficient to recover the shape 
analytically, our deep neural network approach recovers shape 
by learning the shape and corresponding shading in the limited lighting environment for a sufficient sample.

\begin{figure}[t]
    \centering
    \vspace{-5mm}
\begin{minipage}[b]{.45\linewidth}
    \centering
    \includegraphics[width=1.0 \linewidth]{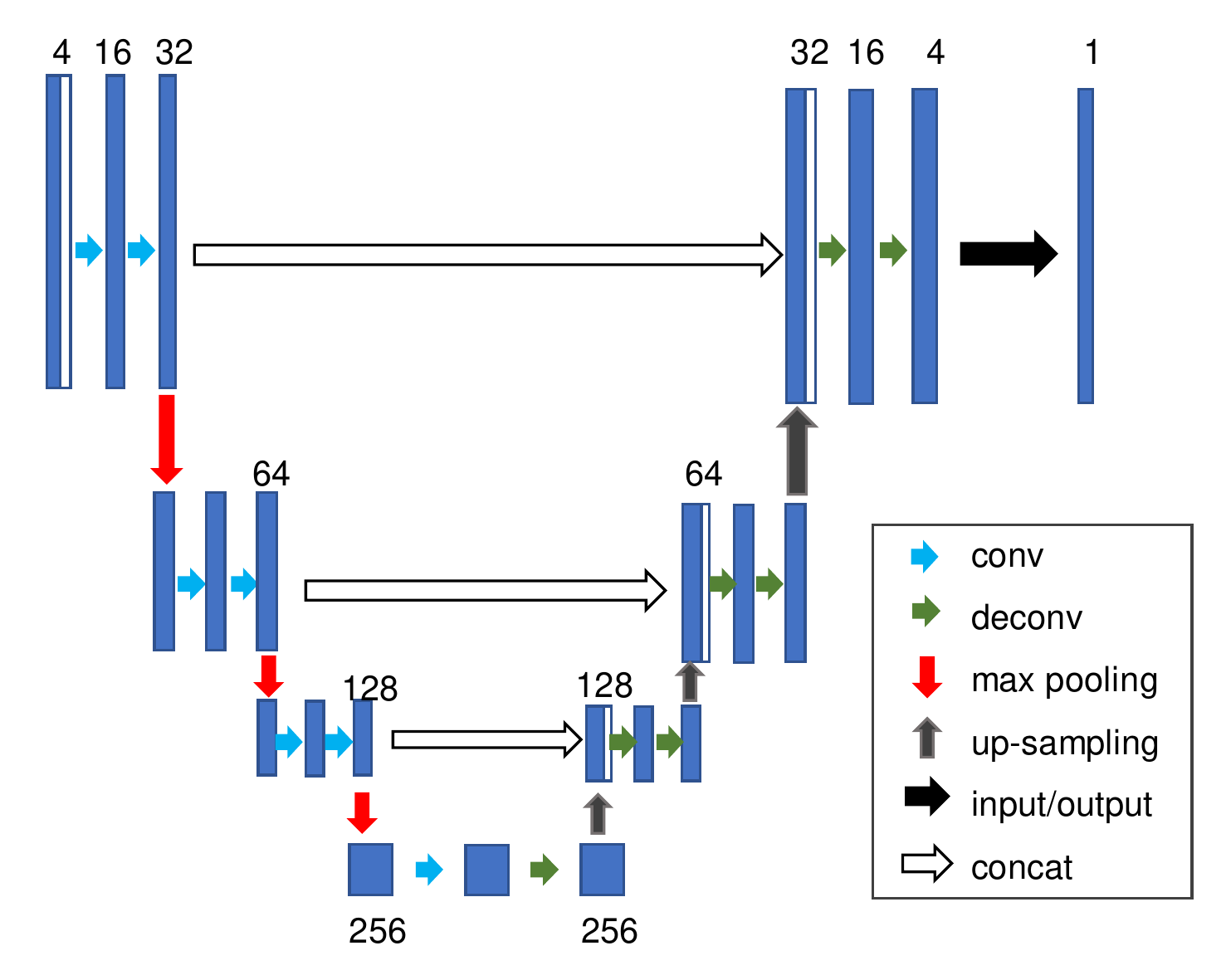}\\
    \vspace{-2mm}
(a)
\end{minipage}
\begin{minipage}[b]{.4\linewidth}
    \centering
    \includegraphics[width=1.0 \linewidth]{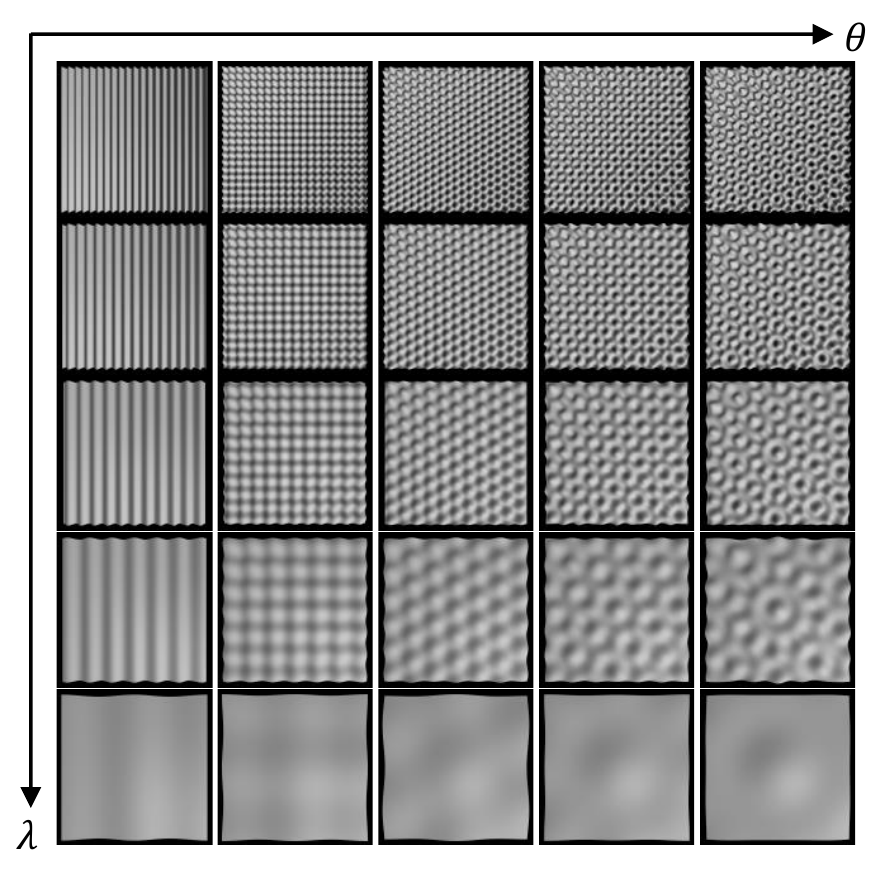}\\
    \vspace{-2mm}
(b)
\end{minipage}
    \vspace{-3mm}
    \caption{(a) The proposed network for shading learning is based on the U-Net structure.
    (b) Examples of shading image of the shape when varying $\theta$ and $\lambda$, which generate many variations of the shape, even for the two parameters.}
    \label{fig:network_structure}
    \vspace{-5mm}
\end{figure}

We utilize CNN to reconstruct consistent shape by taking the shading and low-frequency shape of a local region into account.
U-Net~\cite{Ronneberger2015} structure is employed to retain high-resolution feature of the shading images.
Our network structure is shown in \fref{fig:network_structure}(a).
The proposed network takes three images as input: a shading image, a low-frequency depth image, and a pattern projection image.
Low-frequency depth images are expected to give additional clue to the low-frequency normal direction, and pattern projection images are expected to help in estimating the pixel-by-pixel confidence of low-frequency depth images.
The three images are concatenated in the channel dimension.
The output of the network is the high-frequency shape, which is given by the difference between the low-frequency shape and the ground truth in the training.

After input is converted into 32-dimensional feature by multiple convolutions, low-resolution feature is extracted by pooling layer, then, it is converted into a high-resolution feature by deconvolution.
The shading images and projected images are normalized to a maximum value of 1.
We set the kernel size to 3 and the number of layers to 3.
For loss function, Mean Squared Error (MSE) is computed for the area where the low-frequency depth is available.

\subsection{Training using synthesized data and fine-tuning} \label{sec:synthetic}
A ground truth of high-frequency shape and its shading image are required for the training.
For the robust training, dataset should contain many variations of shape in many poses thus it takes lot of time.
Moreover, high-frequency shapes are difficult to measure.
We propose a domain adaptation method for measuring objects in the real environment by learning a dataset by augmenting it with synthetic images generated by rendering graphics.

To cover wide variation of the shape, one approach is to utilize public 3D shape dataset.
However, even if we use all the existing public high-resolution 3D dataset for our model, it cannot be enough to training the model due to lack of high-frequency shape variations of those dataset.
Instead of using dataset already exists, we generate shapes procedurally to cover many shapes and to make it easier to fine-tune the parameters for the learning.

We generate height map by combining multiple sinusoids:
\begin{equation}
	I(x, y)= \sum_{i=0}^N \alpha_i \cos \left( 2 \pi x^{\prime} \lambda_i + \psi_i \right), x^{\prime} = x \cos \theta_i + y \sin \theta_i,
\end{equation}
where $\alpha$ is coefficient, $\lambda$ is wavelength, and $\psi$ is phase, that are randomly chosen, and $N$ is the number of the basis. 
The parameters are adjusted to obtain a shape close to the real object.
Example of the procedural shapes with regularly varying $\theta$ and $\lambda$ are shown in \fref{fig:network_structure}(b).
The scene is rendered by rasterizing algorithm from the view of the camera which has the same intrinsic matrix in the real measurement.

The synthetic shading image is considered to have different characteristics from the real world, so that models trained using only synthetic data are difficult to utilized to real-world measurements.
In order to support real-world input, the model is trained additionally to fine-tune the weight of the network.
In additional training for fine-tuning, the network is trained with only a few real-world data, and the weights of all layers are updated.

\section{Experiment} \label{sec:experiment}
\subsection{Detail of dataset and evaluation}
We created 600 synthetic data70\% of which were used as training data and the remaining 30\% as validation data.
The parameters of the synthetic shape are N=2, and $\lambda$ is randomized to include the pitch of the cardboards in the real dataset. The object's orientation is randomized within a range of 10 degrees in the pitch and yaw directions and 180 degrees in the roll direction. The direction of the projector is fixed, and its position is randomly shifted within the range where the pattern can be projected onto the object.
We prepared 6 types of real objects taken in the real environment: cardboards with three different high-frequency shapes (A, B, and C) and the other three objects with more complex surface shapes (\shoes, \mesh, and \basket).
The data taken in real environment is divided into training and test dataset.
The training dataset contains 16 data for \cardboard-A, 4 for \cardboard-B, and a few of complex shapes, i.e. 4 for \shoes, 2 for \mesh, and 1 for \basket.
The test dataset contains 12 data of cardboard-B and -C, 4 for \shoes, 6 for \mesh, and 2 for \basket.
In training, the $1200\times1200$ resolution data is divided into $120\times120$ resolution patches and luminance values of images are augmented with a scaling range of 0.5 to 1.5, and is processed in the original resolution for the inference.

For quantitative evaluation, the Root Mean Square Error (RMSE) of the output depth to ground-truth (GT) is calculated.
Since we use shading to estimate the shape of the object, scale and indeterminacy of convexity (known as bas-relief ambiguity~\cite{belhumeur1999bas}) can be a problem.
To deal with this scale ambiguity, the mean and standard deviation of the output patch is adjusted to match ones of GT patch.
Patch size for this computation is $49\times49$ so that one period of the high frequency shape can be fully included.

\subsection{Evaluation on different training dataset}
In this experiment, we compare three models trained with different dataset, the model trained with synthesized dataset, one trained with real dataset, and fine-tuned model.
The weight of fine-tuned model is updated from the synthesized model using real dataset.
The synthesized model is trained for 400 epochs,  real data model for 50, and fine-tuned model for 50 after 400 epochs on synthesized data.
For each model, the weight at the point of lowest validation loss is used for the evaluation.

\begin{figure}[t]
\vspace{-3mm}
    \centering
    \includegraphics[width=0.65 \linewidth]{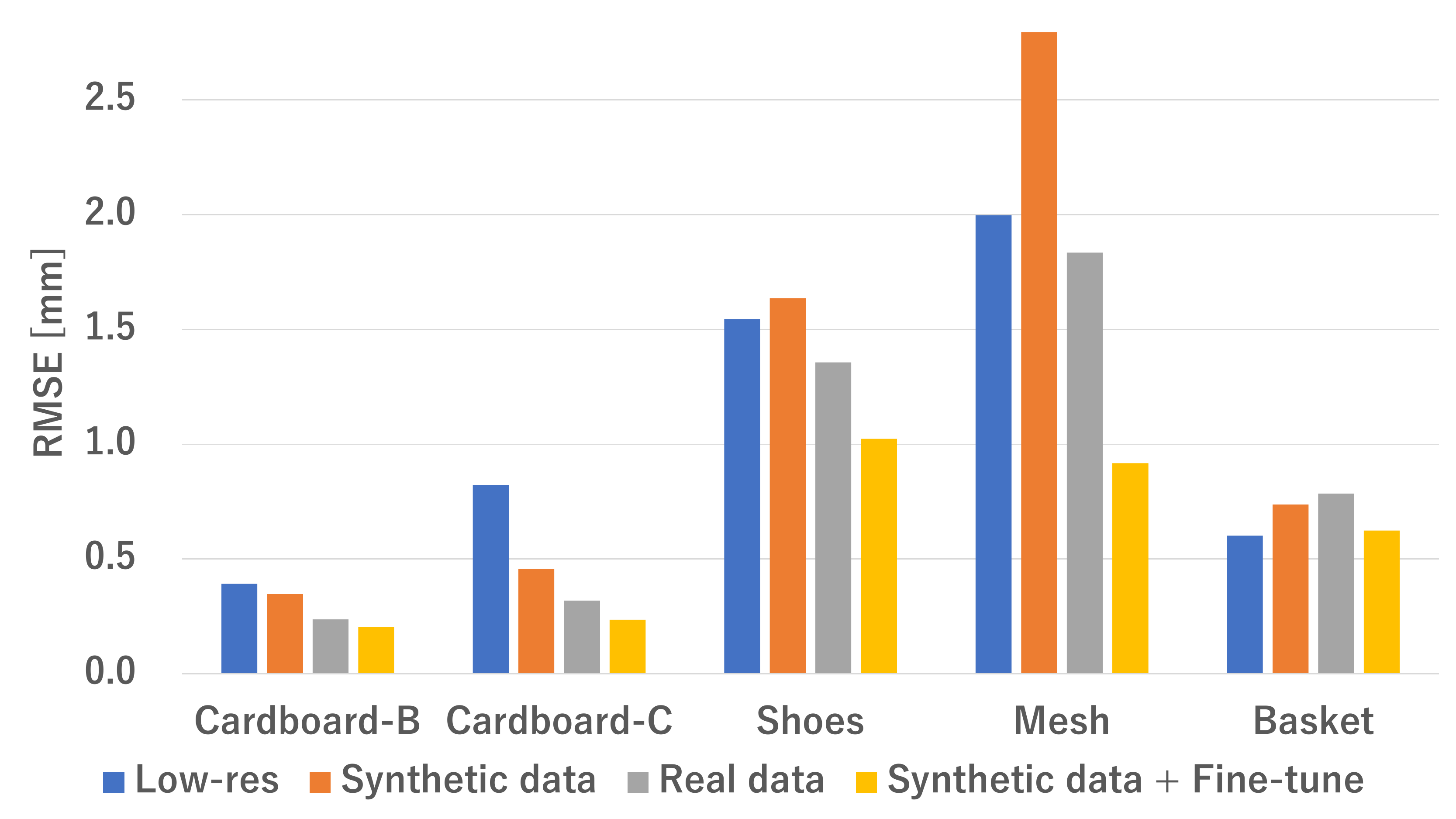}
    \vspace{-5mm}
    \caption{Comparison of errors on training models.}
    \label{fig:result_compare_data}
    \vspace{-5mm}
\end{figure}

\begin{figure}[t]
\vspace{2mm}
    \centering
\hspace{-0.2cm}
\begin{minipage}[b]{0.77\linewidth}
    \centering
    \includegraphics[width=1.\linewidth]{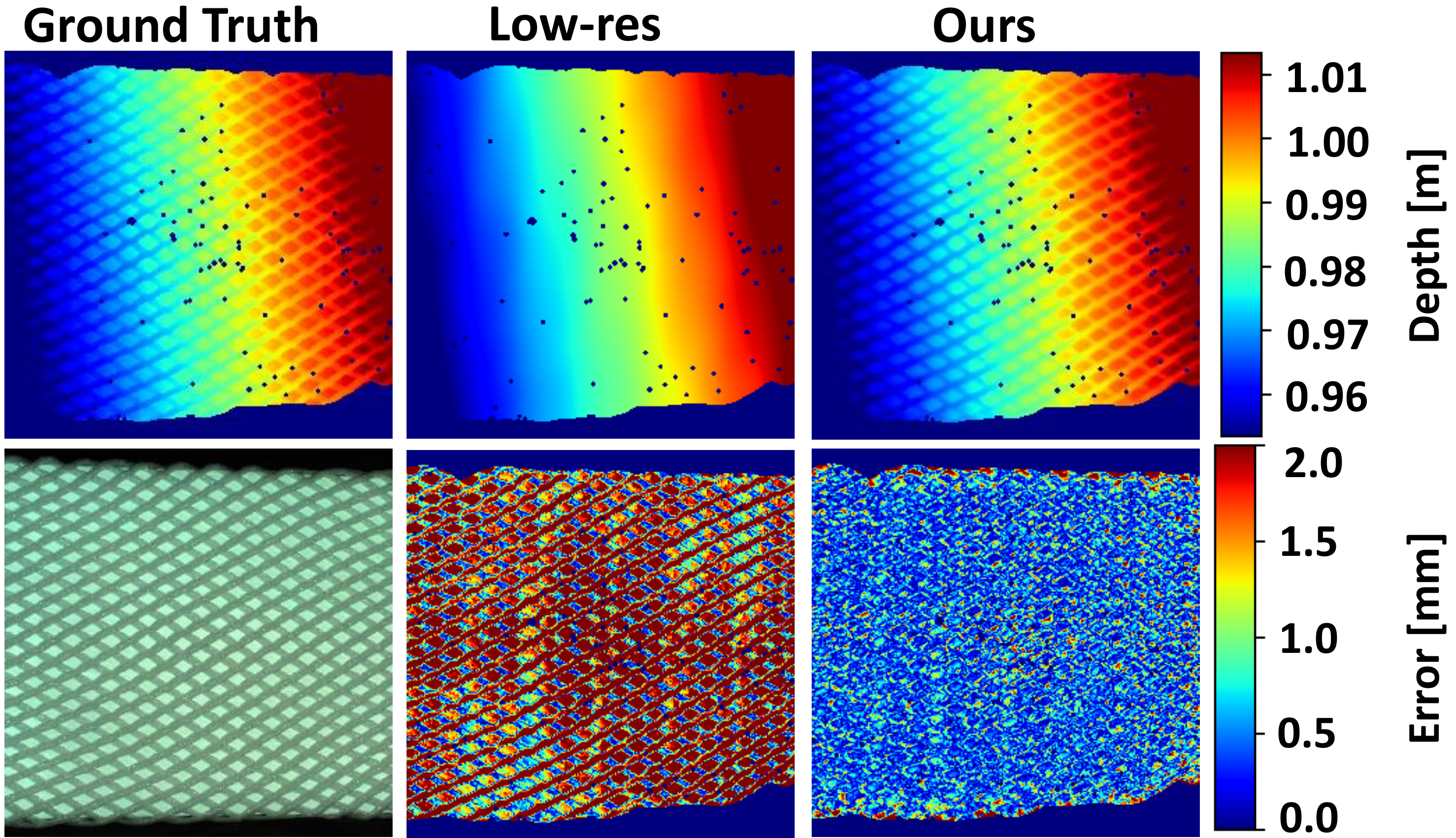}\\
(a)
\end{minipage}
\hspace{-0.2cm}
\begin{minipage}[b]{0.23\linewidth}
    \centering
    \includegraphics[width=1.0\linewidth]{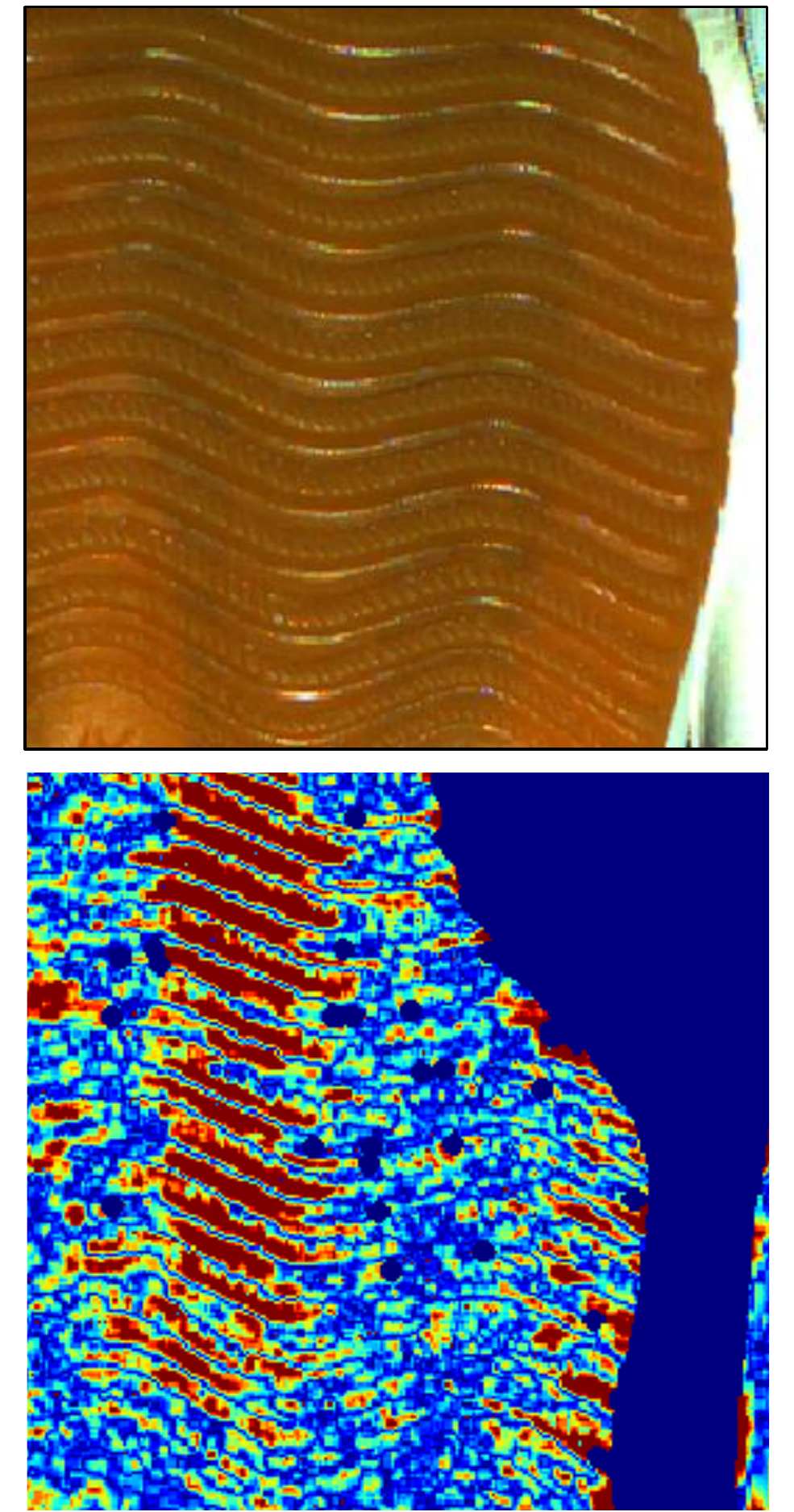}\\
(b)
\end{minipage}
    \vspace{-3mm}
    \caption{(a) Result of \mesh. The top row shows the depth maps, and the bottom row shows the shading image and error maps. (b) Failure case due to specular.}
    \vspace{-5mm}
    \label{fig:error_map}
\end{figure}
The RMSE of depth estimation on the test data of the various domain is shown in \fref{fig:result_compare_data}.
First, we evaluate the performance of the synthetic model. 
Since two cardboard data has high-frequency shapes close to the synthetic training data, the RMSEs are smaller than low-res on the cardboards, but are not improved on complex shapes.
We find that the model trained on synthetic data has the ability to estimate shapes with close domains, but not for complex shapes.

Next, we evaluate the effect of the fine-tuning by comparing with the models trained on synthesized data and real data.
On the cardboards, the performance improved for both types of cardboards including -C, which is not included in the training dataset.
There is a slight improvement on the basket, but on the shoes and mesh, the performance improved significantly after fine-tuning.
The basket is considered to be less effective case since only small number of data are included in the dataset, and the height of the bumps is small compared to other complex shapes.
Overall, the fine-tuned model outperforms the real data model, indicating that the training on the synthetic data contributes to accurate estimation on real data.

\begin{figure}[t]
\vspace{-2mm}
    \centering
    \includegraphics[height=0.35
    \linewidth]{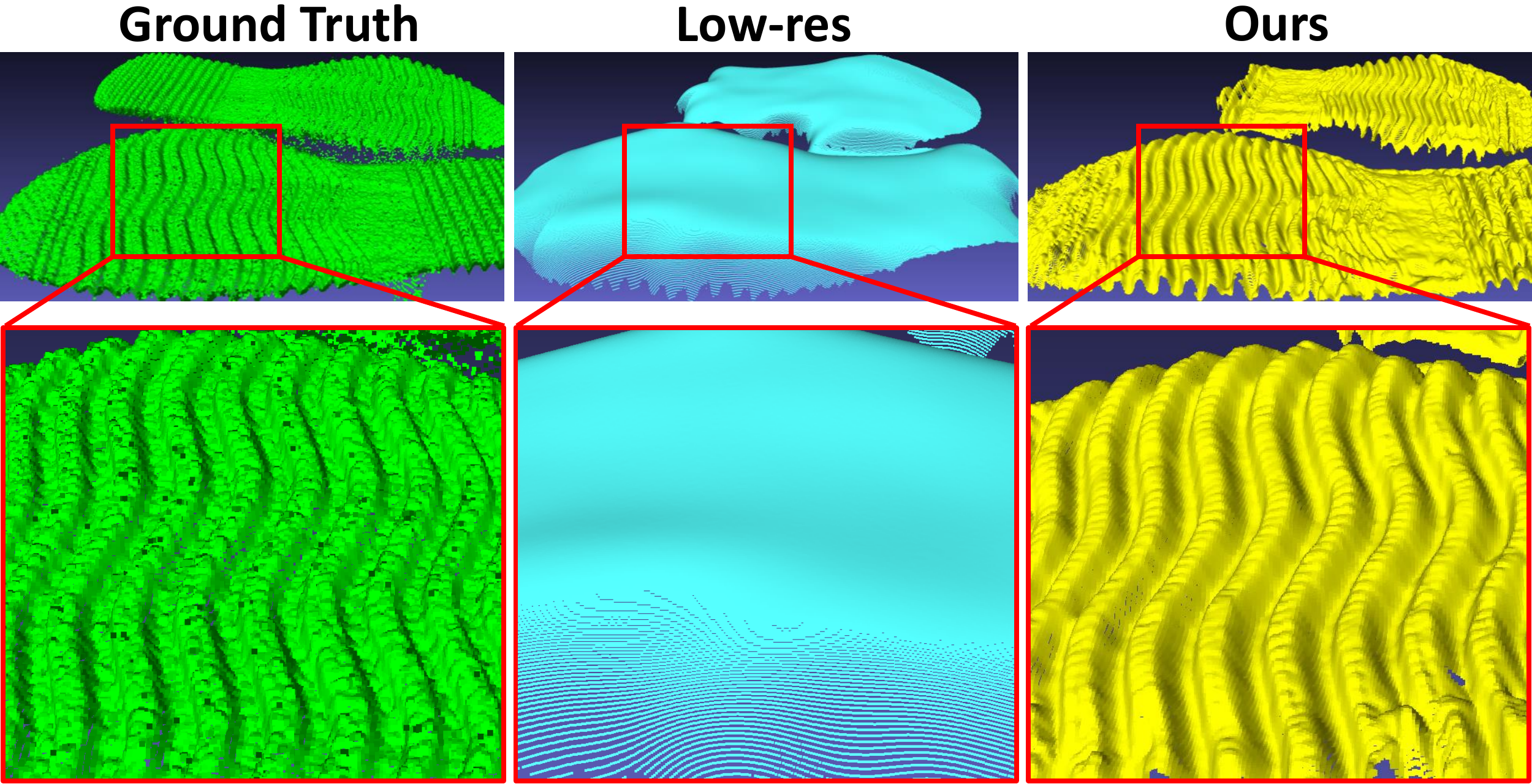}
    \vspace{-3mm}
    \caption{3D reconstruction of \shoes.}
    \vspace{-5mm}
    \label{fig:3d_shoes}
\end{figure}

\fref{fig:error_map} (a) shows the result of the fine-tuned model on mesh test data.
The error map quantitatively shows that our method is able to recover the shape with high accuracy.
However, it may not be possible to estimate correctly due to the specular caused by the material, as shown in \fref{fig:error_map} (b).
Extremely strong specular reflections make it difficult to estimate the correct shape because of saturation and/or excess of dynamic range of the camera.
Looking at the 3D shapes of the shoes test data in \fref{fig:3d_shoes}, we can visually see that the high-frequency shape has been recovered.
We find that our method is able to recover shapes that could not be reconstructed by conventional methods.


\begin{figure}[H]
    \vspace{-1mm}
    \hspace{-0.5cm}
\begin{minipage}[b]{0.23\linewidth}
    \centering
    \includegraphics[width=1.\linewidth]{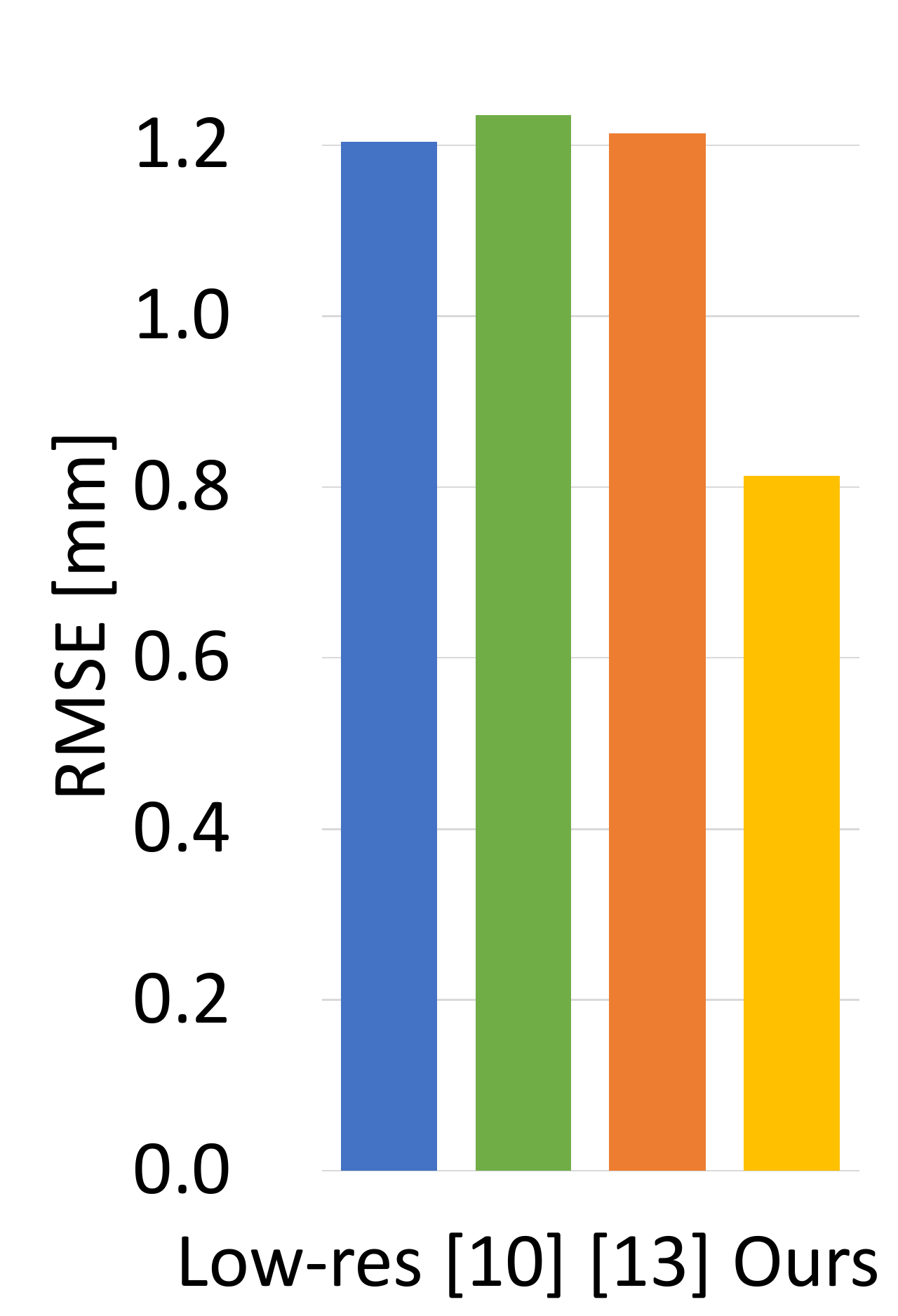}\\
(a)
\end{minipage}
    \hspace{-0.2cm}
\begin{minipage}[b]{.8\linewidth}
    \centering
    \includegraphics[width=1.\linewidth]{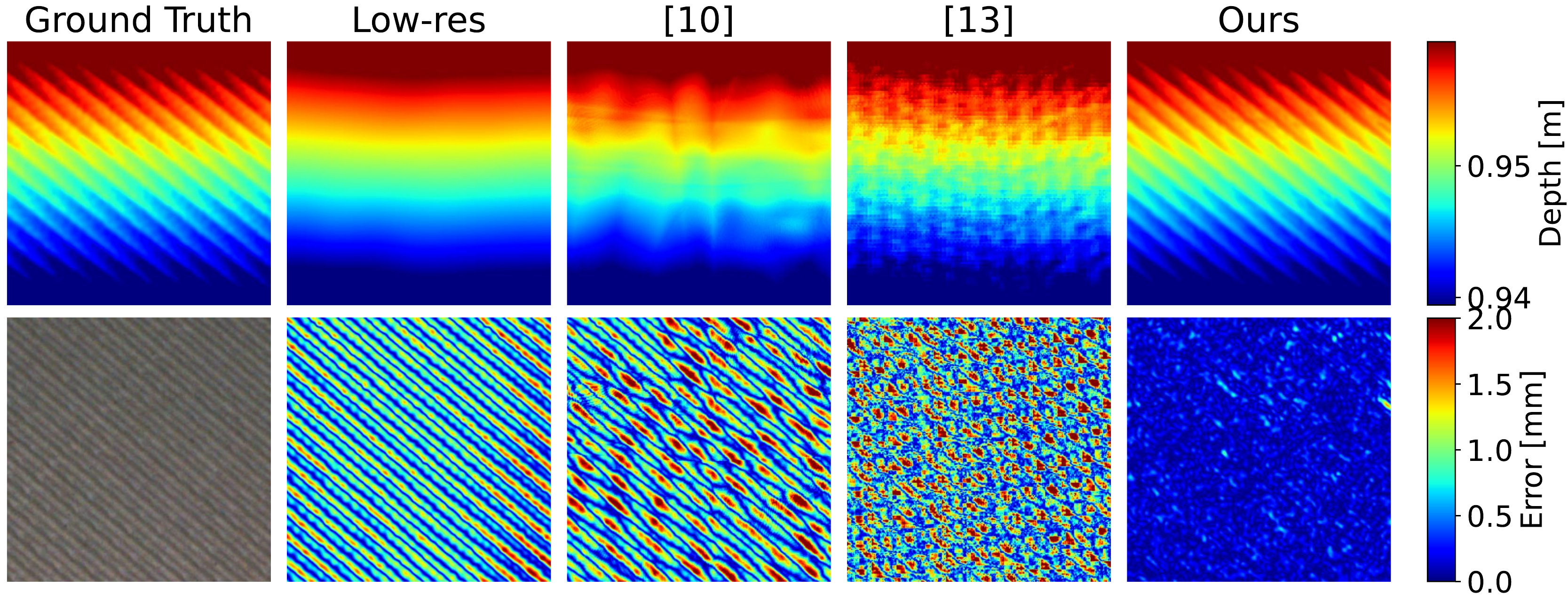}\\
(b)
\end{minipage}
    \vspace{-3mm}
    \caption{Comparison with \cite{Lutio2019} 
    and \cite{Chen_2018_ECCV}. (a) Average RMSE for all methods. (b) Depth and error maps.}
    \vspace{-5mm}
    \label{fig:compare_guide}
\end{figure}


\subsection{Comparison with other methods}
We further compared to the state-of-the-art depth super-resolution 
techniques, such as \cite{Lutio2019} and \cite{Chen_2018_ECCV}. For a clipped $512\times512$ area, the shading image is input as a guide image, and a low-frequency depth image as a low-resolution depth. 
The result is shown in \fref{fig:compare_guide}.
From the figure, the shapes reconstructed by \cite{Lutio2019} and 
\cite{Chen_2018_ECCV} cannot recover high frequency shapes, and thus, averages 
of RMSE
are worse than ours.
One possible reason is non-uniform illumination condition cannot be handled 
correctly by shape super resolution techniques, since the relationship between 
shading and shape varies according to the light position.
On the other hand, our method can make relation between shading and shape 
correctly by using large dataset.

\section{Conclusion} \label{sec:conclusion}
We have proposed a method based on the learning of shading to recover the high-frequency shapes lost in the low-frequency shapes obtained in the measurements.
Since it is difficult to construct a large dataset measured in the real environment, we train the model by a dataset of synthetic images and adapt its domain for real-world images.
It is confirmed that the high-frequency shape of the object in the real environment can be reconstructed.
%
We will investigate methods to eliminate bas-relief ambiguity and to deal with materials that exhibit difficult reflections, such as specular reflections.

\vspace{-0.4cm}
\section*{\centering \large Acknowledgment}
\vspace{-0.3cm}
This work was supported by JSPS/KAKENHI 20H00611, 18K19824, 18H04119, 20K19825 in Japan.
\vspace{-0.4cm}

\bibliographystyle{IEEEbib}
\bibliography{egbib}

\begin{thebibliography}{10}

\bibitem{Horn1974}
Berthold K.~P. Horn,
\newblock {\em Obtaining Shape from Shading Information}, pp. 115--155,
\newblock Winston, P. H. (Ed.), McGraw-Hill, New York, USA, 1974.

\bibitem{Zhang1999}
Ruo Zhang, Ping-Sing Tsai, James~Edwin Cryer, and Mubarak Shah,
\newblock ``Shape-from-shading: a survey,''
\newblock {\em IEEE transactions on pattern analysis and machine intelligence},
  vol. 21, no. 8, pp. 690--706, 1999.

\bibitem{Woodham1980}
Robert~J. Woodham,
\newblock ``{Photometric Method For Determining Surface Orientation From
  Multiple Images},''
\newblock {\em Optical Engineering}, vol. 19, no. 1, pp. 139 -- 144, 1980.

\bibitem{belhumeur1999bas}
Peter~N Belhumeur, David~J Kriegman, and Alan~L Yuille,
\newblock ``The bas-relief ambiguity,''
\newblock {\em International journal of computer vision}, vol. 35, no. 1, pp.
  33--44, 1999.

\bibitem{Yuille1993}
Alan~L. Yuille,
\newblock ``{Impossible Shaded Images},''
\newblock {\em IEEE Transactions on Pattern Analysis and Machine Intelligence},
  vol. 15, no. 2, 1993.

\bibitem{Barron2015}
J.~T. {Barron} and J.~{Malik},
\newblock ``Shape, illumination, and reflectance from shading,''
\newblock {\em IEEE Transactions on Pattern Analysis and Machine Intelligence},
  vol. 37, no. 8, pp. 1670--1687, 2015.

\bibitem{Yang2018}
Dawei Yang and Jia Deng,
\newblock ``{Shape from Shading Through Shape Evolution},''
\newblock in {\em IEEE Conference on Computer Vision and Pattern Recognition
  (CVPR)}, 2018.

\bibitem{Henderson2019}
Paul Henderson and Vittorio Ferrari,
\newblock ``Learning single-image 3d reconstruction by generative modelling of
  shape, pose and shading,''
\newblock {\em International Journal of Computer Vision}, pp. 1--20, 2019.

\bibitem{Barron2016}
Jonathan~T. Barron and Ben Poole,
\newblock ``The fast bilateral solver,''
\newblock in {\em ECCV}, 2016.

\bibitem{Lutio2019}
Riccardo Lutio, Stefano D'Aronco, and Jan Wegner,
\newblock ``Guided super-resolution as a learned pixel-to-pixel
  transformation,''
\newblock in {\em ICCV}, 2019.

\bibitem{Lu_2015_CVPR}
Jiajun Lu and David Forsyth,
\newblock ``Sparse depth super resolution,''
\newblock in {\em Proceedings of the IEEE Conference on Computer Vision and
  Pattern Recognition (CVPR)}, June 2015.

\bibitem{Hui16}
Tak-Wai Hui, Chen~Change Loy, , and Xiaoou Tang,
\newblock ``Depth map super-resolution by deep multi-scale guidance,''
\newblock in {\em ECCV}, 2016, pp. 353--369.

\bibitem{Chen_2018_ECCV}
Zhao Chen, Vijay Badrinarayanan, Gilad Drozdov, and Andrew Rabinovich,
\newblock ``Estimating depth from rgb and sparse sensing,''
\newblock in {\em Proceedings of the European Conference on Computer Vision
  (ECCV)}, September 2018.

\bibitem{Richter2016}
``{Playing for data: Ground truth from computer games},''
\newblock in {\em Lecture Notes in Computer Science (including subseries
  Lecture Notes in Artificial Intelligence and Lecture Notes in
  Bioinformatics)}, 2016.

\bibitem{Richter2015}
Stephan~R. Richter and Stefan Roth,
\newblock ``{Discriminative shape from shading in uncalibrated illumination},''
\newblock in {\em IEEE Conference on Computer Vision and Pattern Recognition
  (CVPR)}, 2015, vol. 07-12-June-2015.

\bibitem{Ikehata2018}
Satoshi Ikehata,
\newblock ``Cnn-ps: Cnn-based photometric stereo for general non-convex
  surfaces,''
\newblock in {\em ECCV}, 2018, pp. 3--18.

\bibitem{Tzeng2017}
Eric Tzeng, Judy Hoffman, Kate Saenko, and Trevor Darrell,
\newblock ``{Adversarial discriminative domain adaptation},''
\newblock in {\em IEEE Conference on Computer Vision and Pattern Recognition
  (CVPR)}, 2017.

\bibitem{Tsai2018}
Yi~Hsuan Tsai, Wei~Chih Hung, Samuel Schulter, Kihyuk Sohn, Ming~Hsuan Yang,
  and Manmohan Chandraker,
\newblock ``{Learning to Adapt Structured Output Space for Semantic
  Segmentation},''
\newblock in {\em IEEE Conference on Computer Vision and Pattern Recognition
  (CVPR)}, 2018.

\bibitem{Hoffman2018}
Judy Hoffman, Eric Tzeng, Taesung Park, Jun-Yan Zhu, Phillip Isola, Kate
  Saenko, Alexei Efros, and Trevor Darrell,
\newblock ``{C}y{CADA}: Cycle-consistent adversarial domain adaptation,''
\newblock in {\em Proceedings of the 35th International Conference on Machine
  Learning}, 2018, pp. 1989--1998.

\bibitem{Sun2016}
Baochen Sun and Kate Saenko,
\newblock ``{Deep CORAL: Correlation alignment for deep domain adaptation},''
\newblock in {\em Lecture Notes in Computer Science (including subseries
  Lecture Notes in Artificial Intelligence and Lecture Notes in
  Bioinformatics)}, 2016.

\bibitem{Furukawa2016}
Ryo Furukawa, Hiroki Morinaga, Shinji Sanomura, Shinji Tanaka, Shigeto Yoshida,
  and Hiroshi Kawasaki,
\newblock ``Shape acquisition and registration for 3d endoscope based on grid
  pattern projection,''
\newblock in {\em ECCV}, 2016.

\bibitem{Ronneberger2015}
Olaf Ronneberger, Philipp Fischer, and Thomas Brox,
\newblock ``U-net: Convolutional networks for biomedical image segmentation,''
\newblock in {\em MICCAI}, 2015, pp. 234--241.

\end{thebibliography}

\end{document}